\documentclass[nonacm, sigconf]{acmart}

\renewcommand\footnotetextcopyrightpermission[1]{} 
\AtBeginDocument{%
	\providecommand\BibTeX{{%
			\normalfont B\kern-0.5em{\scshape i\kern-0.25em b}\kern-0.8em\TeX}}}

\usepackage{latexsym}

\usepackage{url}

\newcommand\BibTeX{B\textsc{ib}\TeX}

\usepackage{latexsym}

\usepackage{url}

\usepackage{subfigure}
\usepackage{amsmath,epsfig}
\usepackage{graphicx}
\usepackage{url}
\usepackage{enumitem}
\usepackage{amsfonts}

\usepackage{algorithm}

\usepackage{amsmath}
\usepackage[noend]{algpseudocode}
\usepackage{mathtools}

\newcommand{\SAVE}[1]{}

\usepackage{mdwlist}  
\usepackage{multirow}  
\usepackage{amsmath} 
\usepackage{relsize} 

\usepackage{float}
\usepackage{flushend}

\usepackage{hyperref}
\makeatletter
\newcommand{\printfnsymbol}[1]{%
	\textsuperscript{\@fnsymbol{#1}}%
}
\makeatother

\usepackage{graphicx}
\makeatletter
\newcommand*\bigcdot{\mathpalette\bigcdot@{.5}}
\newcommand*\bigcdot@[2]{\mathbin{\vcenter{\hbox{\scalebox{#2}{$\m@th#1\bullet$}}}}}
\makeatother

\usepackage{tikz}

\renewcommand{\textrightarrow}{$\rightarrow$}

\usepackage[english]{babel}
\newcommand\HH{
	\global\let\savedtextbullet\textbullet
	\gdef\textbullet{%
		\par\noindent\savedtextbullet\global\let\textbullet\savedtextbullet
	}%
}

\setcopyright{none}

\begin{document}

\title{A Practical Framework for Relation Extraction with Noisy Labels Based on Doubly Transitional Loss}

\author{Shanchan Wu*}
\affiliation{%
	\institution{Alibaba Group (U.S.) Inc.}
	\city{Sunnyvale, CA}
	\country{USA}
}
\email{shanchan.wu@alibaba-inc.com}

\author{Kai Fan*}
\affiliation{%
	\institution{Alibaba Group (U.S.) Inc.}
	\city{Sunnyvale, CA}
	\country{USA}
}
\email{k.fan@alibaba-inc.com}

\thanks{* equal contribution}

\date{}


\begin{abstract}

Either human annotation or rule based automatic labeling is an effective method to augment data for relation extraction. 
However, the inevitable wrong labeling problem for example by distant supervision may deteriorate the performance of many existing methods.
To address this issue, we introduce a practical end-to-end deep learning framework, including a standard feature extractor and a novel noisy classifier with our proposed doubly transitional mechanism. 
One transition is basically parameterized by a non-linear transformation between hidden layers that implicitly represents the conversion between the true and noisy labels, and it can be readily optimized together with other model parameters. 
Another is an explicit probability transition matrix that captures the direct conversion between labels but needs to be derived from an EM algorithm. 
We conduct experiments on the NYT dataset and SemEval 2018 Task 7. 
The empirical results show comparable or better performance over state-of-the-art methods.

\end{abstract}

\maketitle

\section{Introduction}

\begin{figure}[t]
	\centering
	\includegraphics[width=0.95\columnwidth]{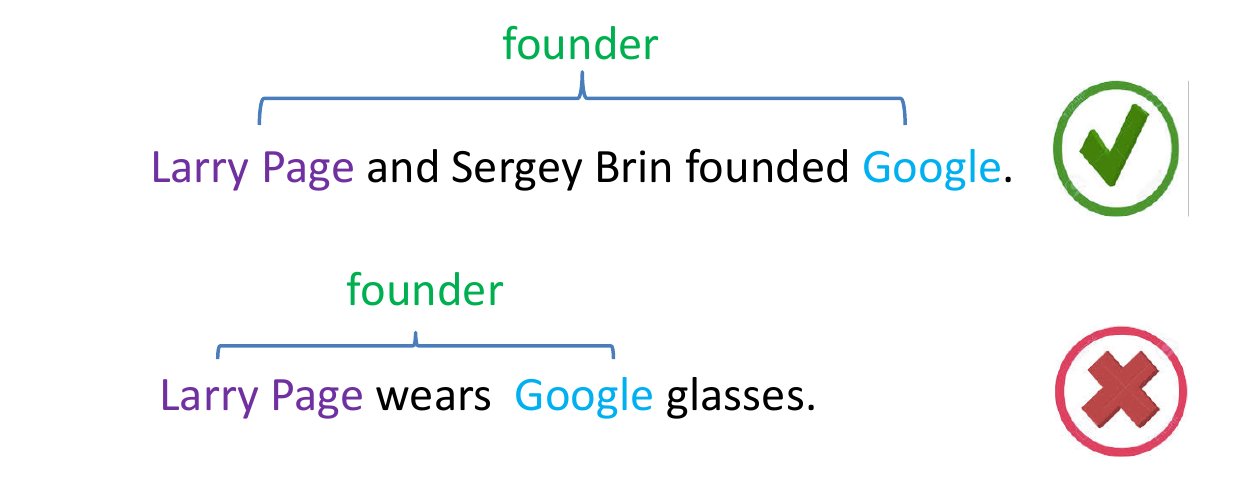}
	\caption{The annotation in KBs does not hold in a specific sentence.
	} 
	\label{fig:def_distance}
	\vspace*{-0.4cm}
\end{figure}

Relation extraction (RE) is the task to predict the semantic relation for a pair of entities
in a text sequence. As the training data is difficult to obtain, many data annotation strategies are used
to alleviate the work load of domain experts, such as crowd-sourcing or distant supervision. 
For an example of distant supervision, a pair of entities in a text sequence is labeled with a relation type
based on the given knowledge bases (KBs). 
In KBs a triplet $(e_1, r, e_2)$ represents a pair of entities $e_1$ and $e_2$ with relation type $r$.
In general, the labeling rule assumes that each sentence containing $e_1$ and $e_2$ will always express the same relation type $r$. 
In this way, the result usually suffers from the noisy labeling problem, such as the example in Figure~\ref{fig:def_distance}. 

To alleviate the impact of the noisy data caused by the annotation, various solutions have been proposed. 
One of the previous works \cite{Shingo_ACL_2012} tries to remove noisy sentences from the training data. 
However, it is not scalable, since it heavily relies on manual rules. 
%
Some other works directly use models to reduce noise rather than filtering out noisy data.
For example, 
\citeauthor{Riedel_ECML_2010} make the \textit{at-least-one} assumption 
that
if two entities participate in a relation, at least one sentence that
mentions these two entities might express that relation.
\citeauthor{Lin_ACL_2016}
propose a sentence-level selective attention mechanism
over a sentence bag with the same entity pair, gaining significant improvement over other previous methods.
Although both the \textit{at-least-one} assumption and the selective attention mechanism
can significantly reduce noise, they are not able to handle the 
situation that given entity pair $(e_1,e_2)$, none of the sentences in the sentence bag expresses the semantic relation $(e_1,r,e_2)$ decided by KBs. 
%
%
%
%
\citeauthor{Wu_AAAI_2019} propose a model with neural noise converter and conditional optimal selector. 
They regard the relation label decided by KBs as a noisy label for the corresponding group of sentences and build a neural noise converter to connect the distribution between true labels and the noisy labels. 
The final prediction based on the conditional optimal selector can outperform the selective attention mechanism.
However, exactly identifying the connection between the true labels and noisy labels
is an `ill-posed' problem. 
For example, given a softmax output, its pre-logits values are not identifiable. 
Thus, \cite{Wu_AAAI_2019} make a very strict constraint on the label transition probabilities, which assumes that only the no-relation type can be mistakenly labeled as other types of relations.      

In this paper, we propose a novel approach to model the distribution of both true labels and noisy labels for relation extraction. 
We first define a probabilistic model with two latent variables, which represents the underlying true label and an indicator to identify whether the true label equals to the noisy label. 
From the Bayesian viewpoint, we develop an Expectation-Maximization (EM) algorithm to optimize an explicit classification loss function. 
The EM algorithm characterizes the connection between the distribution of true labels and noisy labels, and the loss function is derived from the negative log-likelihood of noisy data points by integrating out latent variables.
In addition, in the line of the research of \cite{Wu_AAAI_2019}, we adopt an Implicit Classification Loss as well. 
However, the transition we design is based on the invertible Planar Flow \cite{rezende2015variational}, and the condition we proposed is more robust. 
Collaboratively optimizing the two loss functions can practically achieve better performance.   

For sentence feature extraction and representation, 
we take advantage of pre-trained language models.
Language model pre-training has been shown to be effective
for improving many natural language processing
tasks \cite{Dai_NIPS_2015,Peters_arxiv_2017,OpenAI_2018_tech,Ruder_ACL_2018,bert_Jacobv_corr_bert_2018}. 
For example, both convolutional neural networks
(CNN) \cite{Santos_ACL_2015,Nguyen_NAACL_2015,Zeng_emnlp_2015,Adel_naacl_2016,Wang_ACL_2016} and recurrent neural networks (RNN) \cite{Cai_ACL_2016,Zhou_ACL_2016} have been successfully applied to 
relation extraction. 
The recently proposed language model BERT \cite{bert_Jacobv_corr_bert_2018} 
is extraordinarily remarkable.
We build our sentence representation upon the BERT model to obtain a relational BERT, by adding other necessary components incorporating relational information. 

In summary, our contributions are in four folds: 
\textbf{(1)} We propose an innovative approach with doubly transitional loss to effectively handle the noisy data in relation extraction. 
\textbf{(2)} We adopt the pretrained language model for relation feature extraction. 
\textbf{(3)} We achieve the new state-of-the-art in the distant supervision task for NYT dataset.

\section{Related Work}  
\label{related}

In recent years, deep neural networks have been successfully applied to
relation extraction, typically based on CNN or RNN \cite{Santos_ACL_2015,Nguyen_NAACL_2015, Zeng_emnlp_2015, Adel_naacl_2016, Wang_ACL_2016, Cai_ACL_2016, Zhou_ACL_2016}.
When dealing with noisy labels for relation extraction, some recent approaches have been proposed to reduce the impact of wrong labels by cleaning the noisy training data, such as \cite{Shingo_ACL_2012}.
Some other works directly rely on models to reduce the impact of the noisy data during the training process,
such as \cite{Riedel_ECML_2010,Hoffmann_ACL_2011,Surdeanu_emnlp_2012,Zeng_emnlp_2015}. 
\citeauthor{Lin_ACL_2016} propose an instance-level selective attention model for distantly supervised relation extraction. 
\citeauthor{Wu_AAAI_2019} build a model with neural noise converter and conditional optimal selector to PCNN (a variant of convolutional neural network) for distantly supervised relation extraction. 
Some other NLP related works with noise model include \cite{Fang_CONLL_2016} and \cite{Luo_ACL_2017}.
In the domain of computer vision, researchers also propose many robust strategies including bootstrapping mechanism \cite{Reed_CoRR_2014}, linear noise layer \cite{Sukhbaatar_ICLR_2015} and amortized transition matrix \cite{Misra_CVPR_2016}. 

In general, dealing with noisy labels is important in machine learning
especially when the training data is large. The proposed methods to relieve
the noisy label problems can roughly be classified into three categories \cite{Xiang_2018_ieee}. The first category is to design classification models
to be robust to label noise with some mechanisms like robust loss functions \cite{Beigman_2009_ACL}. The second category \cite{Wilson_ICML_1997} makes effort to 
identify the wrong labeling instances for the purpose of improving
the quality of training data.  
The third category \cite{Lawrence_ICML_2001} aims to model the  distribution of noisy labels during
the training process. By this way, the information of noisy labels
is used for training.

Recently, as large scale deep learning has become popular, training models with noisy labeled data
draws a lot of attention. The reason comes from the fact that deep learning
typically relies on large training data and it is expensive to obtain
sufficient numbers of accurately labeled data. 
For deep learning methods, \citeauthor{Chiyuan_2017_ICLR} \cite{Chiyuan_2017_ICLR} show that a deep network
with large enough capacity can memorize the labels of the training
set even in the circumstances that they are randomly generated. Hence,
general deep networks are particularly susceptible to noisy labels.
\citeauthor{Mnih_2012_ICML} \cite{Mnih_2012_ICML} propose two robust loss functions for aerial images with noisy labels. The limitation of their model
is that it can only be applicable for binary classification.
\citeauthor{Sukhbaatar_ICLR_2015} \cite{Sukhbaatar_ICLR_2015}
propose a model to try to learn the noise distribution. They build a transition
matrix which represents the conversion probability from true labels to
noisy labels. The transition matrix is constructed by adding a 
constrained linear ``noise'' layer on top of the softmax layer,
and it is learned by back-propagating the cross-entropy loss through
the top layer down into the base model.
\citeauthor{Reed_CoRR_2014} \cite{Reed_CoRR_2014} utilize a simple
mechanism to handle noisy labels by incorporating a notion of perceptual
consistency into the usual prediction objective. 
They combine the training
labels and the predictions from the current model to generate
training targets. 
\citeauthor{GhoshKS_AAAI_2017}  \cite{GhoshKS_AAAI_2017} 
investigate the robustness
of different loss functions, such as the mean squared loss,
mean absolute loss and cross entropy loss.
\citeauthor{ZhangS_NIPS_2018} \cite{ZhangS_NIPS_2018} 
combine
advantages of the mean absolute loss and cross entropy loss
to obtain a better loss function.
None of these previous works have combined the explicit loss and
the implicit loss as us to tackle the noisy labeling problems.

\section{Problem Formulation}

Before we introduce our relation extraction system for noisy labels, we first formally define the problem in a probabilistic model, and then we concentrate on the details of each module of the proposed approach. 
Given a set of sentences $\{s_1, s_2, \cdots, s_N\}$, with each containing a corresponding pair of entities $(e_1, e_2)$, the purpose of this task is to predict the relation type $y$ for each entity pair. 
The traditional problem is a supervised learning task by fitting a mapping based on the training dataset, i.e., search a mapping $f: (s, e_1, e_2) \rightarrow y$. 
However, we mainly explore the practical scenario when the labels of training data are possibly contaminated, in other words, the revealed relation type $\hat{y}$ may not correspond to the underlying correct $y$.

The noisy training dataset is usually denoted as a tuple of four elements $\left(s^i, e_1^i, e_2^i, \hat{y}^i\right)_{i=1}^N$. 
The challenge posed in this problem is that the ultimate purpose remains the same as if the data is clean, which still predicts the label as accurate as possible. 
In order to build the connection between the noisy and true labels, we propose a probabilistic framework as follows. 
\begin{align}\label{eq:init}
p(\hat{y}|s, e_1, e_2) &= \int_y p(\hat{y}|y, s, e_1, e_2)p(y|s, e_1, e_2) \mathrm{d}y \\
&= \sum_y p(\hat{y}|y, s, e_1, e_2)p(y|s, e_1, e_2) \nonumber
\end{align}
where we substitute the summation for integration in the second equation, because the relation type $y$ is a discrete variable. 
In addition, the true relation type is a latent variable that we need to make inference. Thus, $p(y|s, e_1, e_2)$ is what we are virtually interested in.


\begin{figure*}[tb]
\centering
\includegraphics[width=1.0\linewidth]{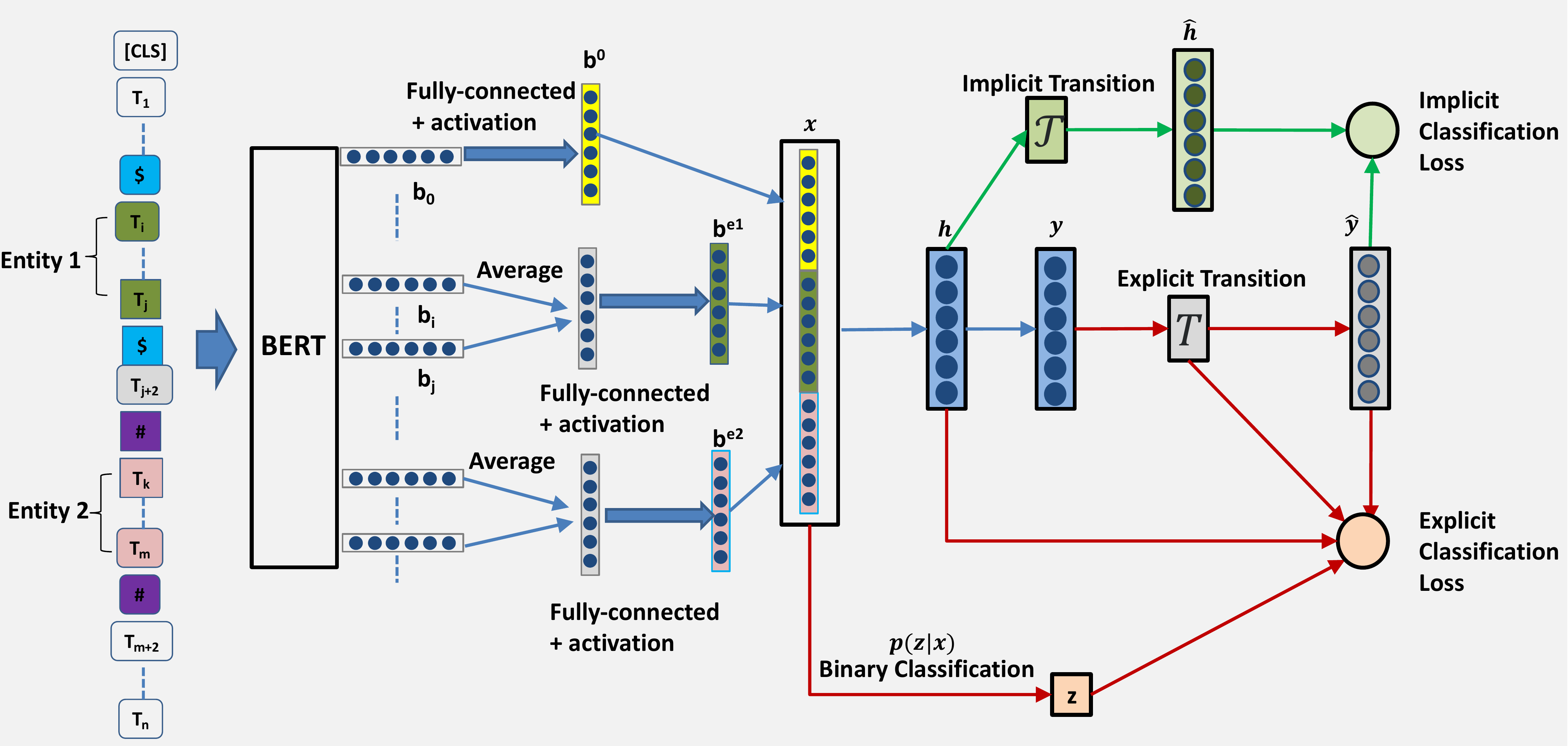}
\caption{The overall model architecture (best viewed in color). The left side shows the sentence feature extractor by a modified BERT language model. The example sentence is eventually encoded to three hidden vectors $b^0,b^{e_1},b^{e_2}$. The right side illustrates the computational graph of our two loss functions, where the green flow shows the implicit classification loss and the red flow shows the explicit classification loss. Note that the output layer $y$ shown as blue is our final prediction layer.} 
\label{fig:model_arch}
\end{figure*}


\section{Methodology}  
\label{sec:methods}

Figure \ref{fig:model_arch} shows the overall neural network architecture of our model. 
It primarily consists of two parts: Sentence Feature Extractor and Noisy Label Classifier.  
Sentence Feature Extractor is used to process a given sentence with a pair of target entities and further obtain a vector representation.
Since we can merely access the noisy labels, it is theoretically forbidden to straightforwardly utilize them to build a classifier. 
Instead, after constructing the vector representation of the sentence, we employ a noisy classifier $p(\hat{y}|y, s, e_1, e_2)$ with our proposed doubly transitional module, which either converts the hidden state of true labels to the hidden state of the noisy labels or directly converts classification result from the true labels to other labels by following the well designed probability distribution. 
Our experimental results demonstrate that the two transitions can potentially achieve mutual improvement in training the underlying true label classifier $p(y|s, e_1, e_2)$.

\subsection{Sentence Feature Extractor}  
 
Recent work BERT \cite{bert_Jacobv_corr_bert_2018} designs a pre-trained language model by multi-layer bidirectional Transformer encoder \cite{Vaswani_NIPS_2017}, achieving the state-of-the-art in many NLP tasks. 
In this section, we will briefly introduce the modifications we made on top of BERT model such that it can readily adapt to a relation extraction task.

\subsubsection{BERT for Relation Extraction}

The left part of Figure~\ref{fig:model_arch} illustrates the major architecture of our sentence feature extractor. 
First, we follow the convention of BERT to insert a special token `[CLS]' at the beginning of each sequence input. 
In our case, each sentence $s$ in principle possibly contains two target entities $e_1$ and $e_2$. 
To make the BERT module capture the location information of the two entities, we suggest inserting two different special tokens `\$' and `\#' at the beginning and the end of both entities. 

Given a sentence $s$ with entity $e_1$ and $e_2$, after the aforementioned preprocessing, a sequence of latent representations is output for each token from the pre-trained BERT model. 
We denote the $b_0$ as the hidden state corresponding to `[CLS]', which is supposed to govern the information of the entire sentence. 
Suppose the subsequence vectors $(b_i,\cdots,b_j)$ are the output hidden states for entity $e_1$, we can obtain a fix-sized summarized vector representation by applying an averaging operation, following by a fully connected layer. 
This process can be formalized as $b^{e_1} = \textbf{FC}\left(\frac{1}{j-i+1} \sum_{t=i}^{j} b_t  \right)$, where \textbf{FC} means a fully connected layer with \texttt{tanh} activation function. 
Similarly, $b^{e_2}$ can be calculated for entity $e_2$ using the same \textbf{FC} layer, i.e., sharing parameters for the weights and bias. 
Since the averaging operation is not necessary for $b_0$, we simply calculate it by $b^0=\textbf{FC}(b_0)$ with a different fully connected layer. 
Note that we also apply dropout before each fully connected layer during training. 
Eventually, we have our extracted sentence feature ready by concatenating $b^0,b^{e_1},b^{e_2}$.


\subsection{Explicit Label Transition} 

We first extract the sentence feature as the concatenation of the three vector representations from relational BERT, denoted as $x=\textbf{Concat}(b^0, b^{e_1}, b^{e_2})$. 
Thus, our interested $p(y|s, e_1, e_2)$ can be simplified as $p(y|x)$. 
However, we can barely access the true label $y\in\{1,...,K\}$ of the training data. 
Instead of directly training a classifier by pretending all noisy labels to be true, we propose a probability transition matrix $T\in \mathcal{R}^{K\times K}$ to capture to what extent a true label will become an incorrect one. 
In this way, the true label $y$ can be treated as the latent variable to be inferred.

We employ a probabilistic framework to describe our approach with the introduction of another latent variable $z$ to indicate whether the noisy label is correct or not. 
Particularly, we denote $z=1$ as the case that the noisy label is exactly the true label while 0 means the noisy label can be any other incorrect label. 
In accordance to the Bayesian Rule, the overall probabilistic latent variable model $p(y,z|\hat{y}, x, \theta)$ can be factorized as the follows.
\begin{equation}
p(y,z|\hat{y}, x, \theta) \propto p(\hat{y}|y,z)p(y|x,\theta)p(z|x,\theta) \label{eq:prob}\\
\end{equation}
where $\theta$ is the parameters of the neural networks in our model, and the transition matrix related term is
\begin{equation}
p(\hat{y}|y,z) = 
\begin{cases}
    \hat{y}^\top I y, & \quad \text{if } z = 1\\
    \hat{y}^\top T y, & \quad \text{if } z = 0
\end{cases}
\end{equation}
Note that \textbf{i)} $I$ is the identity matrix, \textbf{ii)} to make it mathematically concrete, $\hat{y}$ and $y$ are both one-hot vectors, \textbf{iii)} the probability transition matrix $T$ has non-negative elements and zero-valued diagonal, and its column summation is 1. 
The $T_{ik}$ means the probability that the true label $k$ is annotated as another label $i$. 
Since we have no information on the latent indicator $z$ as well, we propose to use an Expectation-Maximization (EM) framework to infer the distribution of the latent variables $y$ and $z$. 

\subsubsection{E-step and M-step}

Since the missing value $z$ follows a discrete distribution, it is possible to further factorize the probabilistic model by iterating a finite set of a fixed number of values. 
In our case, $z$ can only take binary values, allowing us to integrate out the latent variable in E-step. 
According to Eq~(\ref{eq:prob}), we can write the joint distribution for the latent variables in details. 
%
\begin{gather}
	\begin{aligned}\label{eq:e_step}
		p(y=\hat{y},z=1|\hat{y},x) &= \frac{p(y=\hat{y}|x)p(z=1|x)}{C} \\
		p(y=\hat{y},z=0|\hat{y},x) &= p(y=e_k,z=1|\hat{y}, x) = 0 \\
		p(y=\mathbf{e}_k,z=0|\hat{y},x) &= \frac{  p(y=\mathbf{e}_k|x)p(z=0|x) T_{\hat{i}k}   }{C} \\
		C = p(y=\hat{y}|x)p(z=1|x) &+ \sum_{k\neq \hat{i},k=1}^K p(y=\mathbf{e}_k|x)p(z=0|x)  T_{\hat{i}k}   
\end{aligned}
\raisetag{20pt}
\end{gather}
where $\mathbf{e}_k \neq \hat{y}$ denotes the one hot vector where the $k$-th element is 1, 
and $\hat{i}$ is the index of non-zero element of $\hat{y}$.
$C$ is used for normalization in Bayesian Rule.
We simplify the notation by removing the model parameter $\theta$ and the probability transition matrix $T$. 

The $Q(T|T^t)$ function in EM algorithm is the expected value of the log likelihood function of $T$ with respect to the current conditional distribution of the latent variable $z$ given the observed evidence $x$ and the current estimates of the transition matrix $T^t$. 
We can write the $Q$ function over the training dataset as follows to conclude the E-step.
\begin{equation} 
\begin{split}
Q(T|T^t) =& \sum_{n=1}^N\sum_{y_n=1}^K\sum_{z_n=0}^1 \bigg\{ p(y_n,z_n|\hat{y}_n,x_n, T^t) \\
          &  \times \log p(y_n,z_n|\hat{y}_n,x_n, T) \bigg\} \\
\text{s.t.   } & \forall k=1,...,K, \sum_{i\neq k,i=1}^K T_{ik}=1   \notag
\end{split}
\end{equation}
The subsequent M-step requires to maximize the quantity as the following optimization problem $T^{(t+1)}=\arg\max_T Q(T|T^t)$. 
Fortunately, the solution can be analytically derived by Lagrange multiplier method. 
Due to space limitation, we sketch the derivation of the solution of $T$ as follows.

From the definition of $Q(T|T^t)$, we can further write it as 
\begin{gather}
	\begin{aligned}
Q(T|T^t) =& \sum_{n=1}^N\bigg\{p(y=\hat{y}_n,z=1|\hat{y}_n,x_n,T^t) 
 \times \log p(y=\hat{y}_n|x_n)p(z=1|x_n) \\
 + \sum_{k=1,k\neq\hat{i}}^K &p(y=e_k,z=0|\hat{y}_n,x_n,T^t) 
\times \log T_{\hat{i}k} p(y=e_k|x_n)p(z=0|x_n) \bigg\} \\
= \sum_{n=1}^N &\sum_{k=1,k\neq\hat{i}}^K p(y=e_k,z=0|\hat{y}_n,x_n,T^t) \log T_{\hat{i}k} + \text{Constant} \\
\text{ s.t. } & \forall k=1,...,K, \sum_{i=1,i\neq k}^K T_{ik}=1
\end{aligned}
\raisetag{20pt}
\end{gather}
where $\hat{i}$ is the index of non-zero value for one-hot vector $\hat{y}$, and the Constant only depends on model parameter $\theta$ and the $T^t$ at previous iteration, but does not depend on any element of $T$. 
Thus, with Lagrange multipliers method, we consider the new objective
\begin{equation}
\begin{split}
\tilde{Q}(T|T^t) = &\sum_{n=1}^N \sum_{k=1,k\neq\hat{i}}^K p(y=e_k,z=0|\hat{y}_n,x_n,T^t) \log T_{\hat{i}k} \\
&+ \sum_{k=1}^K \lambda_k \left(1-\sum_{i=1,i\neq k}^K T_{ik}\right)
\end{split}
\end{equation}

After making the derivative of $\tilde{Q}$ with respective to each element of $T$, such that $\frac{\partial \tilde{Q}(T|T^t)}{\partial T_{\hat{i}k}} = 0$
\begin{align}
& \frac{\sum_{n \in \{\hat{y}_n = e_{\hat{i}}\}} p(y=e_k,z=0|\hat{y}_n,x_n,T^t)}{T_{\hat{i}k}} - \lambda_k = 0 \\
\Leftrightarrow & T_{\hat{i}k} = \frac{\sum_{n \in \{\hat{y}_n = e_{\hat{i}}\}} p(y=e_k,z=0|\hat{y}_n,x_n,T^t)}{\lambda_k}
\end{align}
Together with the property $\sum_{\hat{i}=1,\hat{i}\neq k}^K T_{\hat{i}k}=1$, we have
\begin{equation}\label{eq:t_update_v1}
\begin{split}
& \lambda_k = \sum_{\hat{i}=1,\hat{i}\neq k}^K \sum_{n \in \{\hat{y}_n = e_{\hat{i}}\}} p(y=e_k,z=0|\hat{y}_n,x_n,T^t) \\
\Leftrightarrow & T_{\hat{i}k} = \frac{\sum_{n \in \{\hat{y}_n = e_{\hat{i}}\}} p(y=e_k,z=0|\hat{y}_n,x_n,T^t)}{\sum_{\hat{i}=1,\hat{i}\neq k}^K \sum_{n \in \{\hat{y}_n = e_{\hat{i}}\}} p(y=e_k,z=0|\hat{y}_n,x_n,T^t)}
\end{split}
\end{equation}
%

%
Based on the meaning of $\hat{i}$, we can rewrite Eq~(\ref{eq:t_update_v1}) as:
\begin{equation}\label{eq:t_update_v2}
T_{ik} = \frac{\sum\limits_{n\in\{\hat{y}_n=\mathbf{e}_i\}} p(y=\mathbf{e}_k,z=0|\hat{y}_n=\mathbf{e}_i,x_n)}{\sum\limits_{i\neq k,i=1}^K\sum\limits_{n\in\{\hat{y}_n=\mathbf{e}_i\}} p(y=\mathbf{e}_k,z=0|\hat{y}_n=\mathbf{e}_i,x_n)} 
\end{equation}
where the numerator can be computed in accordance to the third formula in Eq~(\ref{eq:e_step}).  
 
\subsubsection{Explicit Classification Loss}

We use a binary and a softmax classifier networks to parameterize the required probability distributions in the EM algorithm, $p(z|x,\theta)$ and $p(y|x,\theta)$ respectively, as shown in the right side of Figure~\ref{fig:model_arch}. 
Thus, for a single observation $(x, \hat{y})$, we usually want to minimize the negative log-likelihood $-\log p(\hat{y}|x)$ of this data point. 
However, since it is intractable in our case, we instead propose to optimize its upper bound as our proposed explicit classification loss function. 
We also sketch the derivation of our loss function.

In general, we want to optimize the negative log-likelihood as our loss function in the following formulation. 
%
\begin{gather}
	\begin{aligned}
		p(\hat{y}|x) =& \int p(\hat{y}|y,z)p(y|x)p(z|x)\mathrm{d}y\mathrm{d}z \\
		=& \sum_{k=1}^K \sum_{z=0}^1 p(\hat{y}|y,z)p(y=e_k|x)p(z|x) \\
		=& p(z=1|x)\sum_{k=1}^K \hat{y}^\top I e_k p(y=e_k|x) \\
		&+ p(z=0|x)\sum_{k=1}^K \hat{y}^\top T e_k p(y=e_k|x)  \\
		= p(z=1|&x) p(y=\hat{y}|x) + p(z=0|x)\sum_{k=1,k\neq\hat{i}}^K T_{\hat{i}k} p(y=e_k|x) 
\end{aligned}
\raisetag{20pt}
\end{gather}
%
Notice the fact that $p(z=1|x) + p(z=0|x) = 1$, thus we can use the Jensen's inequality to obtain
%
\begin{gather}
	\begin{aligned}
		-\log p(\hat{y}|x) \leq&  - p(z=1|x) \log p(y=\hat{y}|x) \\
		- p(z=0|x) &\log\left( \sum_{k=1,k\neq\hat{i}}^K T_{\hat{i}k} p(y=e_k|x) \right) \\
		= p(z=1|x)&\cdot \textbf{XE}(h,\hat{y}) \\
		- p(z=0|x) &\Bigg\{ \log\left( \sum_{k=1,k\neq\hat{i}}^K \frac{T_{\hat{i}k}}{T_{\hat{i}\cdot}} p(y=e_k|x) \right) + \log T_{\hat{i}\cdot} \Bigg\}
\end{aligned}
\raisetag{20pt}
\end{gather}
%
where $T_{\hat{i}\cdot} = \sum_{k=1,k\neq\hat{i}}^K T_{\hat{i}k}$, then we use the Jensen's inequality again.
%
\begin{gather}
	\begin{aligned}
		-\log p(\hat{y}|x) \leq& p(z=1|x)\cdot \textbf{XE}(h,\hat{y}) \\
		- p(z=0|x)& \Bigg\{ \sum_{k=1,k\neq\hat{i}}^K \frac{T_{\hat{i}k}}{T_{\hat{i}\cdot}} \log p(y=e_k|x) + \log T_{\hat{i}\cdot} \Bigg\} \\
		= p(z=1&|x)\cdot \textbf{XE}(h,\hat{y}) \\
		+ p(z=0&|x) \Bigg\{ \sum_{k=1,k\neq\hat{i}}^K \frac{T_{\hat{i}k}}{T_{\hat{i}\cdot}} \cdot \textbf{XE}(h, e_k) - \log T_{\hat{i}\cdot} \Bigg\}
\end{aligned}
\raisetag{20pt}
\end{gather}
%

We use the upper bound of negative log-likelihood $-\log p(\hat{y}|x)$
as our proposed explicit classification loss function:

%
\begin{equation}
\begin{split}
\mathcal{L}_e(\theta; x, & \hat{y}, T) = p(z=1|x) \textbf{XE}(h, \hat{y}) + p(z=0|x) \\
      &  \times \bigg\{ \sum_{k\neq \hat{i},k=1}^K \frac{T_{\hat{i}k}}{T_{\hat{i}\cdot}}\textbf{XE}(h, \mathbf{e}_k) - \log T_{\hat{i}\cdot} \bigg\}  \notag
\end{split} 
\end{equation}
where 
$T_{\hat{i}\cdot}$ is the $\hat{i}$-th row summation, $h$ is the logits before applying softmax operation for predicting the true label, and \textbf{XE} means the cross-entropy loss with respect to logits. 
Note that \textbf{i)} the latent variable $z$ is agnostic, we integrate out $z$ in the loss function by taking the advantage of possible binary values, \textbf{ii)} only $\theta$ is optimized by gradient descent through $\frac{\partial L_e}{\partial \theta}$, \textbf{iii)} we conduct an efficient \textbf{iterative training} between $\theta$ and $T$, that is to say, we iterate the following two steps: fixing $T$ to optimize the above loss for $J\geq1$ iterations by stochastic gradient descent based optimization and updating $T$ with the proposed EM algorithm. 

\subsection{Implicit Label Transition} 

Instead of directly building connection between $y$ and $\hat{y}$, we alternatively suggest using another transition mechanism to capture the underlying relation between two types of labels. 
In this case, we assume the existence of the probability distribution of noisy labels $p(\hat{y}|x)$ and a ``ghost" logits $\hat{h}$ such that $p(\hat{y}|x)=\text{softmax}(\hat{h})$. 
We naturally have the following total probability equation,  
\begin{equation} \label{eq:p_total}
\text{softmax}(\hat{h}) = \left( p_{z=0}T + p_{z=1} I \right) \text{softmax}(h) .
\end{equation}
Our intuition is that whether we can build an implicit transformation between the two logits rather than the labels. 
However, softmax operation is not invertible, preventing us to use the transition $\hat{h}=Wh$ if no constraint is imposed on $W$. 

We notice the fact that if every single element of the logits changes, the final probability distribution will be completely rescaled.  
Therefore, we use a two-step transformation $\hat{h}=\mathcal{T}(h)$ modified from planar flow \cite{rezende2015variational},
\begin{equation} \label{eq:flow}
\begin{split}
h' &= h + u\textbf{tanh}(w^\top h + \beta) \\
\hat{h} & = h_1'w' + h_{\cdot\backslash1}'
\end{split}
\end{equation}
where $\beta$ is a scalar, $u,w,w'\in\mathcal{R}^K$ are learnable vectors, and $h_1'$indicates first element of $h'$ while $h_{\cdot\backslash1}'$ means a vector by setting the first element of $h'$ as 0. 
The proposed transformation does not theoretically guarantee the identifiability. 
However, the following properties can guarantee it if our optimization can proceed properly. 

\newtheorem{property}{Property}

\begin{property}
Mapping $h\rightarrow h'$ is invertible if $w^\top u \geq -1$ \cite{rezende2015variational}.
\end{property} 

\begin{property} 
Mapping $h'\rightarrow\hat{h}$ cannot guarantee the identifiability, since $w'\leftarrow w' + \Delta w$ will result in the same probabilty after softmax operation. 
It is an identifiable parameterization for Eq~(\ref{eq:p_total}) if $\|w\|_2^2=c$ where $c$ is a constant. (\cite{Wu_AAAI_2019} did not mention this crucial condition.) 
\end{property} 

A simple \textbf{implicit classification loss} can be defined as $\mathcal{L}_i(\theta; x, \hat{y}) = \textbf{XE}(\hat{h},\hat{y})$, where the parameter $\theta$ includes $u,w,w',\beta$ but $T$ in $\mathcal{L}_e$ is no longer needed.
Note that we use a \textbf{pre-training schedule} by fixing $u=\mathbf{0}$ and $w'=\mathbf{e}_1$ to optimize loss $\mathcal{L}_i$ alone for several epochs. 
Then we set all parameters to be trainable, and train the two loss functions $\mathcal{L}_i$ and $\mathcal{L}_e$ alternatively. 
The overall training procedure is summarized in Algorithm \ref{alg:double_loss}.

As comparison, we have two advantages over \cite{Wu_AAAI_2019}. 
First, the first non-linear invertible mapping can make more complicated approximation, since the implicit $h\rightarrow\hat{h}$ in Eq~(\ref{eq:p_total}) is obviously non-linear. 
Though a linear transition on logits space can achieve non-linear transition on probability space, our approach is more flexible. 
The second mapping is actually a linear transformation that perturbs each element of $h'$ with a different scaled $h_1'$. 
Such simple transformations work well in practice, such as NICE \cite{dinh2014nice}. 
However, \citeauthor{Wu_AAAI_2019} 
omit a crucial condition in their proof.
Secondly, our proposed explicit loss function shares the parameters of the true label classifier $p(y|x)$ with the implicit loss, and empirically we find it benefits the overall training. 

One may suggest using more complicated multi-layer neural network to simulate the transition. 
However, we empirically find a fully connected layer as the transition is difficult to optimize, resulting worse performance. 
We highly suspect this is due to the unidentifiable problem.

\begin{algorithm}[t]
\caption{A Practical Framework for Noisy Training with Doubly Transitional Loss}
\begin{algorithmic}[1]
\small
\State Pretrain the parameter in Bert as \cite{bert_Jacobv_corr_bert_2018}.
\State Fix $u=0, w=0, w'=\mathbf{e}_1, \beta=0$
\While{Not Converge}
	\State Optimize $\mathcal{L}_i$ w.r.t. $\theta$ except $T, u, w, w', \beta$.
\EndWhile
\While{Not Converge}
	\For{$j=0;j<J;j++$}
		\State Optimize $\mathcal{L}_e$ w.r.t. $\theta$ except $T, u, w, w', \beta$.
	\EndFor
	\State Compute $T$ with EM algorithm. 
	\State Optimize $\mathcal{L}_i$ w.r.t. $\theta$ except $T$.
\EndWhile
\end{algorithmic} 
\label{alg:double_loss}
\end{algorithm}

\section{Experiments } \label{sec:dataset}

In this section, we demonstrate the effectiveness of our solution by incorporating doubly transitional loss with the pre-trained language model on two datasets, NYT dataset and SemEval 2018 dataset. 
The NYT dataset is a distantly supervised relation extraction task, and the SemEval 2018 dataset is a general relation extraction task.
For distant supervision, we will describe the dataset and the metrics used for our experimental evaluation, and conduct comparison with other baselines and ablation study to further analyze each component of our own approach. 
For the SemEval competition, we briefly describes the dataset and then compares our results with other methods.

\subsection{Distantly Supervised Relation Extraction}
\subsubsection{Dataset and Evaluation Metrics}
\hfill \break
To evaluate our model, we conduct experiments on NYT dataset developed by 
\cite{Riedel_ECML_2010} and has also been used
by \cite{Hoffmann_ACL_2011,Surdeanu_emnlp_2012,Zeng_emnlp_2015,Lin_ACL_2016,Wu_AAAI_2019}.
Similar to previous works, we use the preprocessed version\footnote{\url{https://github.com/thunlp/OpenNRE}} which is made publicly available by Tsinghua NLP Lab.

This dataset was generated from New York Times corpus (NYT) with relations aligned
with Freebase. The dataset is separated into training set and testing set by first dividing
the Freebase related entity pairs into training and testing parts. 
The training data set is then created by aligning the sentences from the corpus of the years 2005-2006 with the training entity pairs, and the testing data set is created by aligning the sentences from the corpus of the year 2007 with the testing entity pairs. 

Totally there are 53 possible relationships including a special relation type `NA' which indicates no relation between the two entities. The training data set contains 570,088 sentences and 
the testing data set contains 172,448 sentences.


Similar to previous works \cite{Mintz_ACL_2009,Lin_ACL_2016,Wu_AAAI_2019},
the evaluation is on the held-out testing data.
The evaluation compares the extracted relations of the entity pairs
from the sentences in the test data set against the Freebase relation data.
It makes the assumption that the inference model
has similar performance in relation instances inside
and outside Freebase. 
We report the precision/recall curves, Precision@N (P@N), and average precision as 
the metrics in our experiments.

\subsubsection{Parameter Settings}
\hfill \break
%
%
%
We use batch size as 32 and maximum sentence length as 128 in all experiments.
Longer sentence will be truncated to the end tokens. 
For any entities being truncated, we set the positions of the corresponding entities to be the end of the sentence after tokenization. 
We use the Adam \cite{kingma2014adam} optimizer with initial learning rate 5e-5 and weight decay 0.01 for $L_2$ model regularization. 
The dropout rate during training is 0.1 for fully connected layers and BERT module (the uncased basic version).

We pretrain our model with Implicit Classification Loss and without Explicit Classification Loss for 2 epochs as Algorithm \ref{alg:double_loss}.
We then fine-tune our model with all trainable parameters.
During the fine-tuning, we use the following strategy for the initialization of $w'$. 
We define a ratio $\epsilon$, and assign $w'_{1} = 1-\epsilon$, and the rest $K-1$ elements to be $\epsilon/ (1-K)$. 
We set $\epsilon=0.1$ in practice.  

When training with explicit classification loss, we first initialize transition matrix $T$ by setting the diagonal elements to be 0 and all other elements to be  $1 / (K-1)$. 
We efficiently update $T$ by calculating the statistics from every 100,000 sentences instead of the whole dataset.

\subsubsection{Predictions}
\hfill \break
The original prediction
output of our system is the probability distribution of relation types for each test
sentence with its corresponding entity pair. We use the following strategy 
to assign relation types to entity pairs after getting the probability distributions of relation types for all sentences.
%
For a pair
of entity, if all sentences containing the entity pair are predicted to be
negative (i.e. where `no-relation' type has the highest probability 
for each sentence), we make `no-relation' prediction
for the entity pair. Otherwise, if any sentence is
predicted to be some positive relation, we make prediction 
based on the sentences with positive relation
predictions, regardless of those sentences predicted to be`no-relation'.
The selection of positive labels for prediction is based 
on their probability values. We pick the positive label
with the maximum probability value among all positive labels as the prediction
of the entity pair.

\subsubsection{Comparison with Baseline Methods in Literature}
\hfill \break
To evaluate our proposed approach, we select several baseline methods for comparison
on the held-out test dataset, including \cite{Mintz_ACL_2009,Hoffmann_ACL_2011,Surdeanu_emnlp_2012,Lin_ACL_2016,Wu_AAAI_2019}. 

The method that \citeauthor{Mintz_ACL_2009} propose is a traditional distant
supervised model \cite{Mintz_ACL_2009}.
%
The method that \citeauthor{Hoffmann_ACL_2011} propose is a
probabilistic, graphical model for multi-instance
learning that can handle overlapping relations \cite{Hoffmann_ACL_2011} .
%
The method that \citeauthor{Surdeanu_emnlp_2012} propose is a method that models
both multiple instances and multiple relations \cite{Surdeanu_emnlp_2012}.
%
The method \citeauthor{Lin_ACL_2016} propose is a method
that first represents a sentence with PCNN, and then
uses sentence-level selective attention to model a group of sentences with the same entity pair \cite{Lin_ACL_2016}.
%
The method \citeauthor{Wu_AAAI_2019} propose also represents
a sentence based on the architecture of PCNN 
and then applies a neural noise converter and
a conditional optimal selector to reduce the impact of noised data \cite{Wu_AAAI_2019}.

\begin{figure}[tb]
	\centering
	\includegraphics[width=1.1\linewidth]{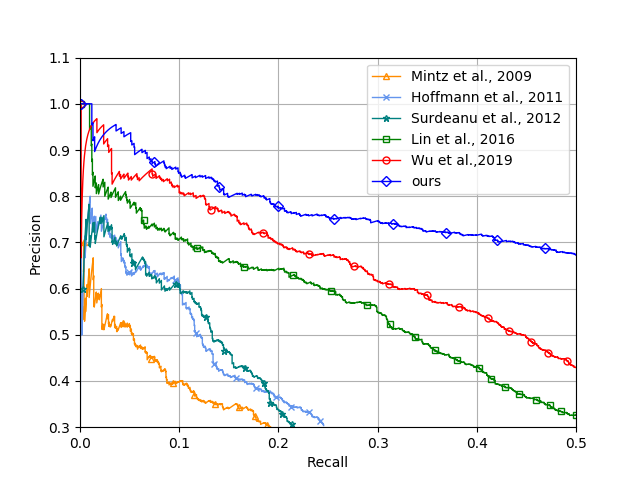}
	\caption{Performance comparison of proposed
                    model and baseline methods. 
	} 
	\label{fig:exp_compare_all}
	\vspace*{-0.2cm}
\end{figure}

Figure \ref{fig:exp_compare_all} shows the precision/recall curves for all methods, 
including ours.  
For all of the baseline methods, we can see that the method in \cite{Lin_ACL_2016} shows much better performance than the methods in 
\cite{Mintz_ACL_2009},  \cite{Hoffmann_ACL_2011} and \cite{Surdeanu_emnlp_2012},
which demonstrates the effectiveness of the sentence-level selective attention. 
The method proposed in \cite{Wu_AAAI_2019} has further improvement than the method 
in \cite{Lin_ACL_2016} and
other baseline methods.
Although the method proposed in \cite{Wu_AAAI_2019} has shown significant improvement over other baselines, our method still gains great improvement over  the method in \cite{Wu_AAAI_2019}. 
Particularly, Table \ref{tab:p_n_noise_vs_pcnnatt} compares the precision@N (P@N) between our model and 
the methods in \cite{Lin_ACL_2016} and \cite{Wu_AAAI_2019}.
Our method achieves the highest values for P@100, P@200, P@300, with mean value of 5.5 higher than the 
method in \cite{Wu_AAAI_2019}, and 14.6 higher than the method in \cite{Lin_ACL_2016}.  
%
 %

\begin{figure}%
\includegraphics[width=0.45\columnwidth]{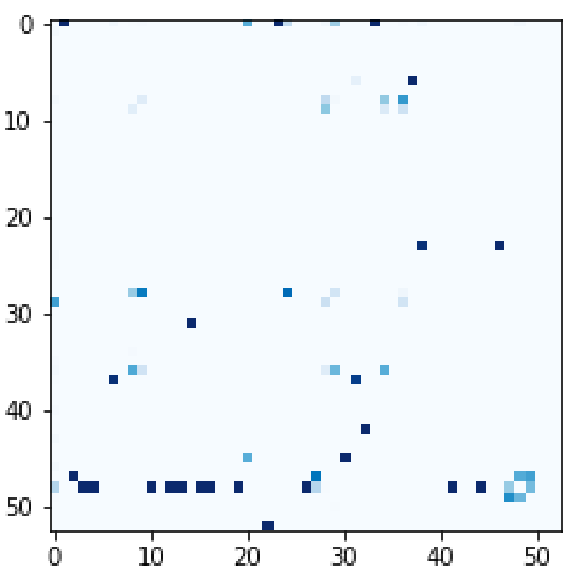} 
\includegraphics[width=0.45\columnwidth]{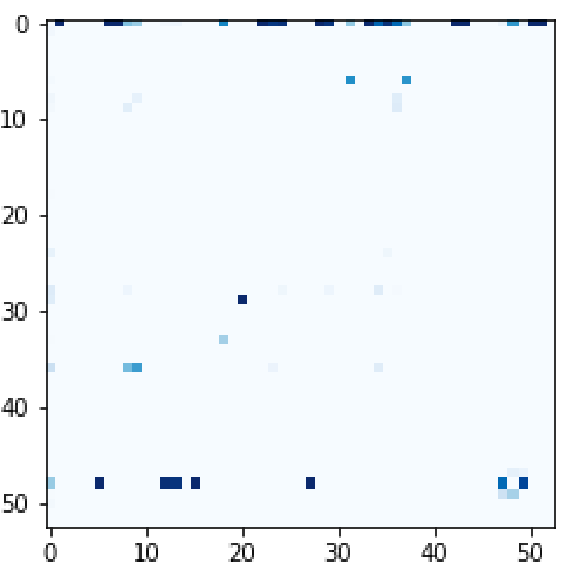}
\caption{Visualization of train/test $T$ matrix .}%
	\label{fig:transition_matrix}%
\end{figure}

%
We further compare the average precision of our method against 
the algorithms in \cite{Lin_ACL_2016} and \cite{Wu_AAAI_2019} in the last column of Table \ref{tab:p_n_noise_vs_pcnnatt},
achieving 21.2 and 28.2 higher average precision scores respectively.  
Comparing to the metric P@100, P@200, P@300, our method has much
more improvement on the metric of average precision. The reason 
can be intuitively summarized as following. 
P@100, P@200, P@300 are corresponding to the precision on the precision recall
curve when the recall is low (roughly in the range of 0.05 to 0.15) in Figure \ref{fig:exp_compare_all}.
Our method has much higher precision improvement when the recall is high than when the
recall is low. For example, from Figure \ref{fig:exp_compare_all}, we can see that 
when the recall is 0.5, the precision of our method is 0.68, 
while the precision of the method in \cite{Wu_AAAI_2019} is 0.43. 
In addition, we also visualize the explicit transitional matrix in the trained model, and the estimated matrix from test dataset in Figure~\ref{fig:transition_matrix}. 
We can see most the elements for the two matrices are similar, demonstrating the derived transitional matrix by the EM algorithm can successfully generalize to test data.
\begin{table}[t]
\centering
\setlength{\tabcolsep}{3pt}
	\caption{Comparison of P@100, P@200, and P@300 and average precision (AP) for relation extraction 
		for our model
		 and the models in  \cite{Lin_ACL_2016} and \cite{Wu_AAAI_2019}
		 }  \label{tab:p_n_noise_vs_pcnnatt}
\begin{tabular}{|c|c|c|c|c|c|c|}
	\hline
	\multicolumn{2}{|c|}{P@N (\%)}  & 100 &200 &300 & Mean & AP (\%) \\
	\hline
	\cline{2-6}
		\hline
	\multicolumn{2}{|c|}{Lin et al., 2016} & 76.2 & 73.1 & 67.4 & 72.2 & 36.5\\
	\hline
	\multicolumn{2}{|c|}{Wu et al., 2019}  & 85.0 & 82.0 & 77.0 & 81.3 & 43.5 \\
	\hline
	\multicolumn{2}{|c|}{ours}  & \textbf{92.0} & \textbf{86.0} & \textbf{82.3} & \textbf{86.8} & \textbf{64.7} \\
	\hline
\end{tabular}

\end{table}

\subsection{Model Analysis}
\subsubsection{Verification of Transitional Loss}
\hfill \break
The method in \cite{Wu_AAAI_2019} has shown to significantly outperform other
previous methods, which is benefited from their proposed Noise-Converter and
Conditional Optimal Selector components.  When comparing with another strong method in \cite{Lin_ACL_2016},
 \citeauthor{Wu_AAAI_2019} in \cite{Wu_AAAI_2019} use PCNN component 
 for sentence representation
 as well as 
\citeauthor{Lin_ACL_2016} in \cite{Lin_ACL_2016}. The only different parts for the two methods 
are the components above the sentence representation, where
 \citeauthor{Wu_AAAI_2019} use Noise-Converter and
 Conditional Optimal Selector and \citeauthor{Lin_ACL_2016} use
 sentence-level selective attention.

To demonstrate  superiority of our method, we further conduct another experiment to 
compare our Transition\_Loss components with the components proposed in  \cite{Wu_AAAI_2019} 
above the same sentence representation. For this purpose, we create a new baseline method, which
uses the same component for sentence representation as ours, and uses Noise-Converter and
Conditional Optimal Selector components as \cite{Wu_AAAI_2019} on top of it. The sentence representation
component is left part till vector $x$ shown in Figure \ref{fig:model_arch}. We label this newly created baseline
method as \textbf{Relation\_BERT+noise\_converter}. Our method is named \textbf{Relation\_BERT+Transition\_Loss}.

\begin{figure}[tb]
	\centering
	\includegraphics[width=1.1\linewidth]{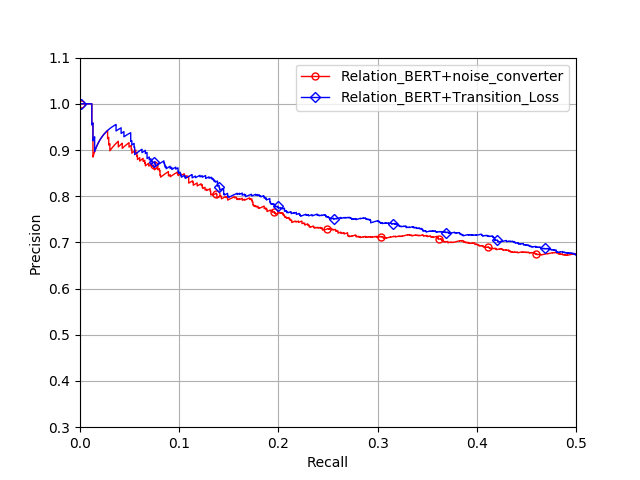}
	\caption{ Performance comparison of our approach and the created baseline method: Relation\_BERT+noise\_converter. 
	} 
	\label{fig:exp_vs_noise_converter}
\end{figure}

Figure \ref{fig:exp_vs_noise_converter} shows the precision/recall curves for this newly created baseline approach Relation\_BERT+noise\_converter
and our method  Relation\_BERT+Transition\_Loss.
We can see that our method shows better performance than the created baseline method.  
Table \ref{tab:noise_convert_vs_transition_loss} compares the precision@N (P@N) and average precision (AP) for the two methods. The comparison of P@1000 of the two methods are shown 
in Table \ref{tab:noise_p1000}. 
Our method achieves the higher values for all these metrics than the baseline method.  

\begin{table}[t]
	\centering
	\setlength{\tabcolsep}{3pt}
	\caption{Comparison of P@100, P@200, and P@300 and average precision (AP) for relation extraction 
		for our model
		and the created baseline Relation\_BERT+noise\_converter
	}  \label{tab:noise_convert_vs_transition_loss}
	\begin{tabular}{|c|c|c|c|c|c|c|}
		\hline
		\multicolumn{2}{|c|}{P@N (\%)}  & 100 &200 &300 & Mean & AP (\%) \\
		\hline
		\cline{2-6}
		\hline
		\multicolumn{2}{|c|}{Relation\_BERT+noise\_converter} & 91.0 & 85.0 & 80.7 & 85.6 & 64.0\\
		\hline
		\multicolumn{2}{|c|}{Relation\_BERT+Transition\_Loss}  & \textbf{92.0} & \textbf{86.0} & \textbf{82.3} & \textbf{86.8} & \textbf{64.7} \\
		\hline
	\end{tabular}
	
\end{table}

\begin{table}[tb]
	\caption{Comparison of P@1000 for two methods: Relation\_BERT+noise\_converte, and  Relation\_BERT+Transition\\\_Loss}
	\label{tab:noise_p1000}
	\begin{tabular}{|c|c|c|c|c|c|}
		\hline
		Methods  & 	P@1000 (\%)  \\	
		\hline
		Relation\_BERT+noise\_converter & 69.8   \\
		\hline
		Relation\_BERT+Transition\_Loss & \textbf{71.5}   \\
		\hline
	\end{tabular}
\end{table}

\subsubsection{Ablation Study}
\hfill \break
A critical point of our approach is that to what extent the prevalent BERT
model can benefit the performance of the relation extraction task, and  
how our proposed doubly transitional loss can even
boost the performance of the model.   

To understand the impact of the components in our approach, we conduct
several more experiments. One experiment is that we use the BERT model alone, 
and the input for the BERT model is sentences containing the target entities
without inserting the special tokens. 
The output hidden vector of `[CLS]'
is then connected to a fully connected layer for classification. 
This method is named \textbf{BERT}.

Another experiment is that we remove the noise related components
in our approach, and keep the rest parts. 
That is to say, we insert the special tokens around the two target entities, and use
the output hidden vectors of the first token and the output hidden tokens
of the two target entities for classification. 
More specifically, in Figure \ref{fig:model_arch}, we remove all components in the right of hidden vector $x$, and add a fully connected layer and softmax layer over $x$ for classification. 
This method is named \textbf{Relation\_BERT}. 
Additionally, our method is named \textbf{Relation\_BERT+Transition\_Loss}.
 
\begin{figure}[tb]
	\centering
	\includegraphics[width=1.1\linewidth]{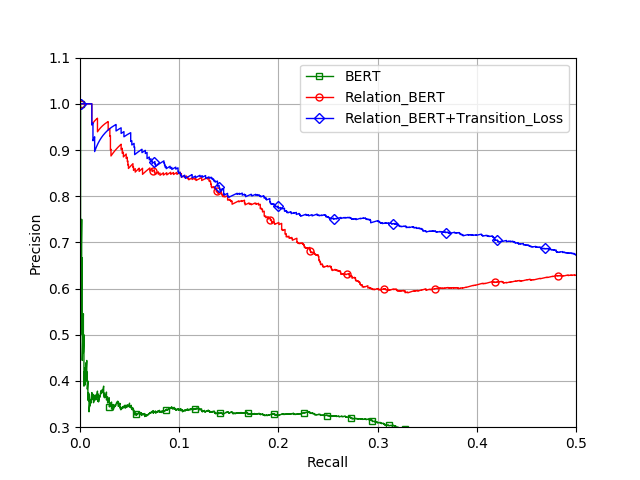}
	\caption{ Performance comparison of our approach and the methods
		with some components removed. 
	} 
	\label{fig:exp_compare_bert}
\end{figure}

Figure \ref{fig:exp_compare_bert} illustrates the precision/recall curves of all comparison methods. From this figure, we can see that the general method
\text{BERT} has the much worse performance than the other two methods. 
Although BERT has shown very strong performance in general text classification \cite{bert_Jacobv_corr_bert_2018}, but without incorporating
the locations of the target entities and their corresponding output hidden vectors,
it does not show any advantage on this task. On the other hand, Relation\_BERT
shows much stronger performance than general BERT, which demonstrates
the effective way that we incorporate the target entity information into
the model.  Furthermore, the approach Relation\_BERT+Transition\_Loss
which incorporate all components in our model architecture gains
significant improvement over Relation\_BERT. We can see that when the 
recall is 0.3, Relation\_BERT+Transition\_Loss has the precision value
of 0.74, while Relation\_BERT has the precision value of
0.6, which shows 14\% precision improvement on this recall point.
The comparison of P@1000 of the three methods are shown 
in Table  \ref{tab:ablation_p1000_ap}.  
Relation\_BERT+Transition\_Loss has the highest
P@1000 value, with 11.6\% absolute improvement over Relation\_BERT,
and 38.8\% absolute improvement over BERT.
Table \ref{tab:ablation_p1000_ap}  
further shows that 
Relation\_BERT+Transition\_Loss has the highest average precision value.

\begin{table}[tb]
	\caption{Comparison of P@1000 and average precision (AP) for three methods: BERT, Relation\_BERT, and Relation\_BERT+Transition\_Loss}
	\label{tab:ablation_p1000_ap}
	\begin{tabular}{|c|c|c|c|c|c|}
		\hline
		Methods  & 	P@1000 (\%) & AP  \\
		\hline
		BERT  & 32.7 & 23.1  \\		
		\hline
		Relation\_BERT & 59.9 & 61.9   \\
		\hline
		Relation\_BERT+Transition\_Loss & \textbf{71.5} & 64.7   \\
		\hline
	\end{tabular}
	\vspace*{-0.4cm}
\end{table}

%
%

We also conduct two more experiments. One only uses the explicit classification loss, 
and the other only uses the implicit classification loss. We report the average precision of
these two methods as long as the method using both loss in 
Table \ref{tab:ablation_loss}.
It shows that when using both loss functions, we have higher average precision. 

\begin{table}[tb]
	\caption{Comparison of average precision (AP) for methods with different loss}
	\label{tab:ablation_loss}
	\begin{tabular}{|c|c|c|c|c|c|}
		\hline
		Methods  & 	AP (\%)  \\
		\hline
		 Explicit Classification Loss  & 64.2  \\		
		\hline
		 Implicit Classification Loss & 64.1   \\
		\hline
		Both Loss & \textbf{64.7}   \\
		\hline
	\end{tabular}

\end{table}
\vspace*{-0.2cm}

\subsection{General Relation Extraction Task}

In this task, we evaluate on the SemEval 2018 Task 7 dataset. 
We focus on subtask 1.1, where the test dataset is composed of clean entities with manual labels. 
Like several other solutions \cite{Jonathan_semeval2018_arxiv}, we combine
the training dataset of subtask 1.1 and the training dataset of subtask 1.2
as the new training dataset. 
The entities in the dataset of subtask 1.2 are automatically annotated and hence contain some noises. 
In this setting, we have the noisy training dataset and evaluate on the clean test dataset. 
Originally there are 6 relation types with 5 of them are asymmetrical and one of them is symmetrical. 
For the 5 asymmetrical relations, we create 5 extra corresponding reversed relations. 
Then we have totally 11 relations. 
For two entities, we consider their relations only when they appear in the same sentence. 
After pre-processing, the training dataset contains 2476 sentences (1248 of them are from subtask 1.1  and
1228 of them are from subtask 1.2), and the test dataset contains 355 sentences. 
This is a much smaller dataset compared with the NYT dataset.   

For the hyper-parameters, we use the similar ones as NYT dataset. 
Since this dataset is much smaller, the following parameters are different. 
In this experiment, we update $T$ for every 16 sentences, and we set batch size to be 16. 
We fine-tune our model for 5 epochs.

\begin{table}[tb]
	\vspace*{-0.4cm}
	\centering
	\setlength{\tabcolsep}{3pt}
	\caption{Comparison with top 5 teams on subtask 1.1 of SemEval 2018 Task 7.}
	{
		\label{tab:comare_semeval}
		\begin{tabular}{|c|c|}
			\hline
			Methods / Teams & F1 score   \\
			\hline
			Talla  & 74.2\\
			\hline
			ClaiRE  & 74.9\\
			\hline
			SIRIUS-LTG-UiO  & 76.7\\
			\hline
			UWNLP  & 78.9\\
			\hline
			ETH\-DS3Lab & 81.7 \\
			\hline
			Ours & 80.7 \\
			\hline
		\end{tabular}
	}

\end{table}

We use the official tool to generate the final macro F1 scores. 
Table \ref{tab:comare_semeval} shows the results of our methods with comparison with the top 5 teams \cite{conf_semeval2018_Gabor} \footnote{\url{https://lipn.univ-paris13.fr/~gabor/semeval2018task7/}} who participated
in the task competition. 
Our method can rank 2nd place with a little bit less than the top 1 team. 
The top 1 team applies a lot of engineering tricks to achieve the high macro F1 score, such as using the
weight of classes (which boosted their F1 for 1.6 points), reversing sentences rather than adding
reversed relation types (which boosted their scores for 2.0 points) \cite{Jonathan_semeval2018_arxiv}. 
We did not apply any engineering tricks. 
The results show that our method is still very promising in general relation extraction task, since the training data is inevitably containing some noisy data.

%

\section{Discussion and Limitation}


In this paper, we construct two types of transitional loss to tackle the relation extraction problem, which shows significant improvement over the competitive baseline methods. 
Optimizing the two losses together also outperforms using only one type of the transitional loss. 
We partially attribute this improvement to the alternative training. 
Switching the two types of loss during optimization potentially helps to escape from local minimal, because the local optimal of one loss function is not necessarily the local optimal of another one. 
Since the direction of gradient descent is characterized by the local landscape around the current optimized parameters, we believe that the two types of transitional loss have a discriminatory gradient at the same point.

A critical point in our work is the relationship of the two losses. 
First, they both share the same sentence representation component and the logits vector used for generating true labels (See Figure 2 for the left parts till vector $h$). 
After back-propagation of gradient descent, both of them will optimize the parameters of the shared components. 
Second, they differ in the way how to model the connection between the noisy (observed) labels and the true labels.   

While the evaluation of our model based on Doubly Transitional Loss shows better performance in practice, it still has two potential limitations. 
First, since the model is based on the pretrained BERT and includes EM optimization process, the computational cost is relatively high and the latency in prediction will be a practical issue for production deployment. 
Second, since our EM algorithm for the explicit loss needs to gather sufficient instances to calculate the sufficient statistics to estimate the desired parameters, the estimation will be in high variance if the size of training data is small.  
This phenomenon can be found in the experiments on the SemEval 2018 Task 7 dataset, which contains less than 3 thousand instances.
Although our solution is very competitive in this dataset, it does not outperform other methods in a large margin. 

When comparing with the baselines, we would address the reason that we do not use the same baselines for both the NYT dataset and the SemEval 2018 Task 7 dataset. 
The latter task, which targets general relation extraction, is a slightly different task from the former one, which aims at distantly supervised relation extraction. 
The baselines from \citeauthor{Lin_ACL_2016} \cite{Lin_ACL_2016} and  \citeauthor{Wu_AAAI_2019} \cite{Wu_AAAI_2019} are particularly suitable for distantly supervised dataset, and hence it is fair enough to compare them with our method on the NYT dataset. 
The methods to solve the two problems should be fundamentally different, but we attempt to unify them with our proposed approach. 
Although the comparison with other top ranked teams in the competition on the SemEval 2018 Task 7 dataset is not strictly fair, our method still achieves an acceptable result.



\section{Conclusions} \label{twitter_sec:conclude}
In this paper, we propose an innovative approach with doubly transitional loss
to effectively handle the noisy data in relation extraction. 
The explicit classification loss is essentially derived from a probabilistic model
with latent variables. The implicit classification loss function represents an end-to-end 
noisy transition framework. 
In the experiments, we validate the effectiveness of each component in our approach, and it also demonstrates comparable or state-of-the-art performance in the relation extraction task with noisy data.

\newpage

\bibliography{mynlp}

\bibliographystyle{acl_natbib}



\end{document}